\documentclass{article}
\usepackage{iclr2027_conference,times}
\iclrfinalcopy

\usepackage{amsmath,amsfonts,bm}

\def\eqref#1{equation~\ref{#1}}

\def\1{\bm{1}}

\DeclareMathAlphabet{\mathsfit}{\encodingdefault}{\sfdefault}{m}{sl}
\SetMathAlphabet{\mathsfit}{bold}{\encodingdefault}{\sfdefault}{bx}{n}

\usepackage{hyperref}
\usepackage{url}
\usepackage{graphicx}
\usepackage{booktabs}
\usepackage{tabularx}
\usepackage{amsmath}
\usepackage{multirow}
\usepackage[font=small,labelfont=bf]{caption}
\usepackage{microtype}
\usepackage{xcolor}
\definecolor{darkblue}{rgb}{0.1,0.2,0.5}
\hypersetup{colorlinks=true,linkcolor=darkblue,citecolor=darkblue,urlcolor=darkblue}
\DeclareMathOperator{\logit}{logit}
\DeclareMathOperator{\FPR}{FPR}

\newcommand{\thalf}{t_{0.5}}
\newcommand{\tthree}{t_{0.3}}
\newcommand{\Hbar}{\bar{H}}

\makeatletter
\def\@maketitle{\vbox{\hsize\textwidth\centering
{\LARGE\sc \@title\par}
\vskip 0.25in
{\bf \@author\par}
\vskip 0.3in minus 0.1in}}
\makeatother

\title{Who Sets the Pace of Judge Hacking?\\Judge Capability versus the Policy--Judge Gap}

\author{Haosen Wang}

\begin{document}
\maketitle
\lhead{Preprint}

\begin{abstract}
Under GRPO against a frozen, reference-free LLM judge, RL exploits the judge's false positives. Does the policy--judge capability gap or the judge itself set how fast and how far? We train 0.5B and 1.5B Qwen2.5 policies on GSM8K against four judges at least as large, and the 1.5B policy also against one smaller judge. In a post-hoc analysis, judge acceptance rose, but no cell of 12 gained more than $0.04$ in seed-mean held-out greedy accuracy. A gold-reward control with the same recipe raised the 1.5B policy's accuracy by $+0.054$ (for the 0.5B policy, gold reward gave no detectable gain either). The depth of judge hacking tracks the judge: the plateau correlates with judge AUC (Spearman $-0.69$ over 12 cells), and within a family the larger judge is fooled less deeply. For the pace, two pre-specified pairs contradict the gap account: the cell with the larger gap was fooled $3.5\times$ sooner at the pre-specified threshold. With 12 cells, however, we cannot formally separate judge AUC from the gap for either endpoint. A blind test of step-0 predictors on three held-out judges was inconclusive. At 7--8B, judges from three families plateaued between $0.37$ and $0.85$.
\end{abstract}

\section{Introduction}

LLM judges and model-based verifiers are used as RL rewards where no hand-written verifier exists \citep{ma2025general,liu2026examining}. A judge that sees a candidate answer scores plausibility, not correctness, and RL can raise its acceptance rate while true accuracy stays flat \citep[``more convincing, not more correct'', in a reasoning-suppressed regime;][]{zhou2026more}. Our logs show this pattern (Section~\ref{sec:gold}). So we do not measure overoptimization in the sense of \citet{gao2023scaling}, where the proxy first helps and then hurts. We measure how fast and how far GRPO exploits a false-positive region that the judge has from step~0.

\begin{figure}[t]
\begin{minipage}[t]{0.44\linewidth}
\vspace{0pt}
\centering
\includegraphics[width=\linewidth]{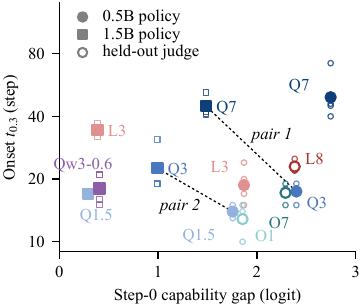}
\caption{\textbf{Cells with similar gaps differ in onset.} Onset $\tthree$ against the step-0 capability gap $\logit(\text{AUC})-\logit(\text{accuracy})$. Small marks: runs; large: AFT means (circle: P0.5; square: P1.5; open: held-out judge). Dashed: pre-specified pairs. Qw3-0.6: judge smaller than the policy (stage 6; 42-run fit).}
\label{fig:main}
\end{minipage}\hfill
\begin{minipage}[t]{0.545\linewidth}
\vspace{0pt}
\centering
\footnotesize
\captionof{table}{\textbf{Pre-specified pairs} (first cell has the larger gap). Gap account: onset differences $\geq0$, plateau $\leq0$. Onsets: 97.5\% (38-run AFT). KL: Welch 95\% (P0.5$\times$Q1.5 lacks sampled KL). $\Hbar$: Welch 95\%.}
\label{tab:pairs}
\setlength{\tabcolsep}{2pt}
\begin{tabular}{lrr}
\toprule
 & Pair 1 & Pair 2 \\
 & {\scriptsize P0.5$\times$Q3 $-$ P1.5$\times$Q7} & {\scriptsize P0.5$\times$Q1.5 $-$ P1.5$\times$Q3} \\
\midrule
Param.\ ratio & 6.3 / 4.9 & 3.1 / 2.0 \\
Capability gap & 2.40 / 1.49 & 1.76 / 0.99 \\
$\ln\thalf$ & $-1.25$ {\scriptsize$[-1.62, -0.88]$} & $-0.64$ {\scriptsize$[-1.09, -0.20]$} \\
$\ln\tthree$ & $-0.94$ {\scriptsize$[-1.17, -0.71]$} & $-0.48$ {\scriptsize$[-0.76, -0.20]$} \\
$\ln\text{KL}$ at $\tthree$ & $-0.84$ {\scriptsize$[-1.20, -0.49]$} & -- \\
$\Hbar$ & $+0.40$ {\scriptsize$[0.24, 0.57]$} & $+0.32$ {\scriptsize$[0.25, 0.39]$} \\
\bottomrule
\end{tabular}
\end{minipage}
\end{figure}

What sets this pace? In the \textbf{gap account}, the capability gap between judge and policy sets it: judge hacking ``worsens precisely when the judge is weaker than the policy'' \citep{kenton2026debate}, and a model of oversight as a function of the capability gap points the same way \citep{engels2025scaling}. In the \textbf{judge account}, the judge itself sets it. Only a \emph{crossed} design, with cells of similar gap and different judges, separates the two. In the main design the judges are 1--16$\times$ larger than the policy and both policies are weak solvers, so there we test the gap account only where the judge is at least as strong as the policy. One added cell pairs the 1.5B policy with a smaller judge (Qwen3-0.6B). This judge solves GSM8K about as well as the policy and comes from another family, so the cell is not a clean test of a judge weaker than the policy. We run GRPO \citep{shao2024deepseekmath} on GSM8K \citep{cobbe2021training}. Every stage had a hashed specification before its runs, and we report all deviations (Appendix~\ref{app:prereg}). Our findings: (1) \textbf{judge-RL did not detectably improve held-out greedy accuracy} (post hoc): no cell of 12 showed a seed-mean gain above $0.04$, judge acceptance rose in all 11 cells where we logged it, and in the four rising cells we checked, including the largest rise, the rise in training-rollout accuracy is mostly sampling sharpening; a gold-reward control with the same recipe gained $+0.054$ $[0.013, 0.096]$ for P1.5, so there the flat accuracy is specific to the judge reward, but only $+0.022$ $[-0.015, 0.059]$ for P0.5, so there it is not (Section~\ref{sec:gold}); (2) \textbf{in our cells with judge $\geq$ policy, the depth of judge hacking (the plateau) tracks the judge}: it correlates with judge AUC (Spearman $-0.69$ $[-0.94, -0.14]$ over 12 cells), and within a family the larger judge is fooled less deeply; for the pace (onset), the two pre-specified pairs contradict the gap account at two onset thresholds (descriptive contrasts), but with 12 cells we cannot formally separate judge AUC from the gap for either endpoint (difference in correlations: plateau $-0.38$ $[-0.95, 0.00]$; onset $0.36$ $[0.00, 0.91]$); the one cell with a smaller judge shows no evidence of a gap effect, but its control plateau is near the ceiling and family and template are confounded (Section~\ref{sec:gap}); (3) \textbf{a blind test of step-0 predictors is inconclusive} and does not favor the screened predictor over AUC (Section~\ref{sec:blind}); (4) \textbf{same-size judges from different families differ widely} (Section~\ref{sec:scale}).

\section{Setup}
\label{sec:setup}

\textbf{Policies and judges.} P0.5 and P1.5 are Qwen2.5-0.5B and -1.5B (base) \citep{qwen2024qwen25} after one epoch of SFT on 2{,}000 GSM8K solutions; step-0 greedy accuracy on 200 held-out test problems is $0.33$ and $0.61$. The main judges are Qwen2.5-1.5B/3B/7B-Instruct (Q1.5, Q3, Q7) and Llama-3.2-3B-Instruct (L3) \citep{meta2024llama32} (26 runs); the held-out judges are Llama-3.1-8B-Instruct (L8) \citep{grattafiori2024llama3} and OLMo-2-0425-1B-Instruct and OLMo-2-1124-7B-Instruct (O1, O7; two releases) \citep{olmo2024olmo2}. Judge names give the family (Q: Qwen2.5, L: Llama, O: OLMo-2) and the size in billions. Stage 6 adds Qwen3-0.6B in non-thinking mode (Qw3-0.6) with P1.5 only (4 runs); it solves $0.585$ of the held-out problems (P1.5: $0.61$). A judge sees the problem and solution, not the reference. The reward is $G_{\text{cal}}=\sigma(aG+b)$ with $G=\logit(\texttt{Yes})-\logit(\texttt{No})$ and $(a,b)$ fitted on step-0 samples.

\textbf{Training.} GRPO on 5{,}473 GSM8K problems, 8 problems $\times$ 8 rollouts per step, learning rate $3\times10^{-6}$, no KL penalty, DAPO token-level loss \citep{yu2025dapo}, no format reward, 150--250 steps. Table~\ref{tab:hparams} in Appendix~\ref{app:details} lists all hyperparameters. Every 5--10 steps we log greedy accuracy on the 200 held-out problems and the KL to the step-0 policy on sampled answers (nats per sequence).

\textbf{Judge prompt (no reference).} ``Below is a math problem and a proposed solution.\textbackslash n\textbackslash nProblem: \{q\}\textbackslash n\textbackslash nSolution: \{s\}\textbackslash n\textbackslash nQuestion: Is the final answer of this solution correct? Answer Yes or No.'' Applied with each model's chat template. The solution is truncated to 1{,}500 characters.

\textbf{Calibration and thresholds.} $G_{\text{cal}}$ is fitted per judge on step-0 samples of both policies (1{,}000 training problems $\times$ 2 samples per policy). $\tau$, $\FPR_0$ and all step-0 predictors use a disjoint set (256 problems $\times$ 8 samples per policy). Predictor intervals use a problem-clustered bootstrap.

\textbf{Sampling and optimization.} Rollouts and step-0 samples use temperature $1.0$ (no top-$k$ or top-$p$) and at most 256 new tokens. Advantages are centred within each group of 8 and divided by the group standard deviation. Each step has 64 rollouts, split into micro-batches with gradient accumulation.

\textbf{Data split and evaluation.} GSM8K shuffled with seed 0: 2{,}000 training problems for SFT, the other 5{,}473 for RL. Held-out evaluation: the first 200 test problems after shuffling, greedy decoding, every 5 steps to step 50 and every 10 steps after that. The greedy answers were saved, and Section~\ref{sec:gold} re-scores them with the lenient label. Sampled KL is estimated on sampled answers to the same problems; the 8 stage-1 runs logged only greedy-answer KL.

\textbf{Endpoints.} The gold label is \emph{lenient} (the endpoint fixed before stage 2): the number after ``\texttt{\#\#\#\#}'', otherwise the last number. A per-answer check of 300 judge-accepted, lenient-wrong answers from the format-drifting cells found 5.7\% parser errors, 12 of the 17 in P1.5$\times$Q7 (Appendix~\ref{app:extra}). Per cell, $\tau$ is fixed at step 0 (true-positive rate $0.80$). FPR is the share of wrong answers with $G_{\text{cal}}\geq\tau$, and the fooled fraction is $h=(\FPR-\FPR_0)/(1-\FPR_0)$ on 5-step windows. Onset $\thalf$ ($\tthree$) is the first step with $h\geq0.5$ ($0.3$) on two consecutive windows. Plateau $\Hbar$ is $h$ over steps 101--150. Onsets use one censored log-normal AFT model over the 38 runs of stages 1--5 (profile-likelihood intervals); the stage-6 cell uses the same model refitted on all 42 runs, which moves the other geometric means by at most $1.3$ steps; plateaus use Welch intervals over seeds (2--4 per cell). The study ran in seven stages (Appendix~\ref{app:prereg}): Q1.5/Q7 (stage 1), Q3 (2), L3 (3), predictor screening (4), a blind test (5), a smaller judge with a sampling-sharpening check (6), and a gold-reward control (7).

\begin{figure}[t]
\centering
\includegraphics[width=\linewidth]{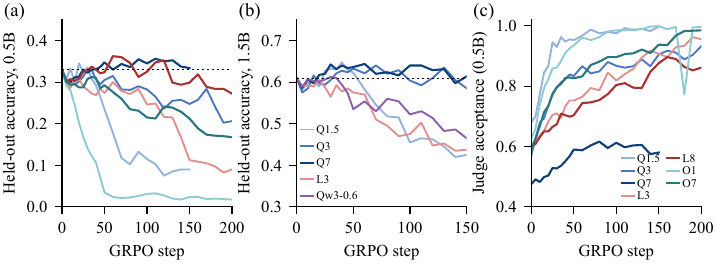}
\caption{\textbf{The judge accepts more answers while held-out accuracy stays flat or falls.} Seed means; 200 held-out problems, greedy, lenient label. (a, b) Accuracy of the 0.5B and 1.5B policies; dashed: step 0. (c) Share of the 0.5B policy's held-out answers that the training judge accepts. Q, L, O = Qwen2.5, Llama-3, OLMo-2; Qw3 = Qwen3; number = judge size in billions. Acceptance was not logged for Qw3-0.6.}
\label{fig:gold}
\end{figure}

\section{Judge-RL did not detectably raise held-out greedy accuracy}
\label{sec:gold}

\begin{figure}[t]
\centering
\includegraphics[width=\linewidth]{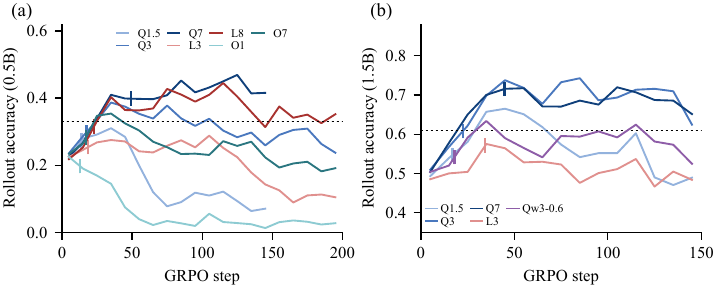}
\caption{\textbf{Training-rollout accuracy rises early in 11 of 12 cells, then falls.} Lenient accuracy of the training rollouts, 10-step windows, seed mean, for the 0.5B (a) and 1.5B (b) policy. Ticks: onset $\tthree$ (AFT geometric mean). Dashed: step-0 \emph{greedy} held-out accuracy. Step-1 sampled accuracy is lower; a check on step-50 checkpoints of four rising cells shows that the rise there is sampling sharpening, with no detectable change in greedy accuracy (Section~\ref{sec:gold}). Qw3-0.6 tick: 42-run fit.}
\label{fig:rollout}
\end{figure}

No cell's seed mean gained more than about one binomial standard error ($0.033$--$0.035$ at $n=200$): the largest seed-mean gain over step~0 was $+0.037$ (P1.5$\times$Q1.5, step 40), and each maximum is taken over 20--25 evaluation steps (Figure~\ref{fig:gold}; Table~\ref{tab:gold}). Single seeds reached up to $+0.085$ (P1.5$\times$Q3), and we ran no equivalence test, so small gains are not excluded. Under Q7 accuracy stayed flat; under weaker judges it fell, for P0.5 to $0.09$ (Q1.5) and $0.02$ (O1) by step 150. Judge acceptance of the same held-out answers rose in every cell where we logged it (all but stage 6), for example from $0.57$ to $0.91$ under Q3. Training-rollout accuracy did rise in 11 of 12 cells (P0.5: from about $0.22$ to $0.29$--$0.47$; Figure~\ref{fig:rollout}). But training-rollout accuracy in steps 1--10 is far below greedy accuracy ($0.22$ vs.\ $0.33$; training problems at temperature 1 vs.\ held-out problems, greedy), and the peaks land near the greedy level. A sharper sampling distribution would produce the same rise, and a pre-registered check (stage 6) classifies both main cells as sharpening. On the 200 held-out problems, the step-50 checkpoints of P0.5$\times$Q3 and P1.5$\times$Q3 gained $+0.13$ $[0.09, 0.17]$ and $+0.23$ $[0.19, 0.27]$ in temperature-1 accuracy over the SFT policy, close to the rise of the training rollouts ($+0.14$ and $+0.23$). Greedy accuracy, measured in the same pass, changed by only $+0.004$ $[-0.054, 0.062]$ and $+0.035$ $[-0.024, 0.094]$ (Table~\ref{tab:sharp}). So in these cells the rise in rollout accuracy is the policy's temperature-1 samples concentrating on its greedy answers, with no detectable change in greedy accuracy (upper bounds $+0.06$ and $+0.09$). In the four rising cells we checked at step 50, including the largest rise (P0.5$\times$Q7: $+0.147$ $[0.113, 0.181]$ at temperature 1, $+0.026$ $[-0.023, 0.075]$ greedy; also P0.5$\times$O7), the rise is mostly sharpening. The check covers step 50 only; the P0.5$\times$Q7 rollout peak is near step 130. Appendix~\ref{app:extra} gives the details, including a failed validity check that does not change the classification.

\textbf{Gold-reward control (stage 7).} We repeated the recipe with the gold label as reward (1 if the lenient label is correct, else 0), 2 seeds and 150 steps per policy, and measured the step-150 checkpoints on 500 held-out problems (Table~\ref{tab:sharp}). For P1.5, greedy accuracy rose by $+0.054$ $[0.013, 0.096]$ (seeds $+0.058$, $+0.050$), so for this policy the flat greedy accuracy under judge reward is specific to the judge reward. For P0.5, the gain was $+0.022$ $[-0.015, 0.059]$ (seeds $-0.008$, $+0.052$): neither judge nor gold reward gave a detectable greedy gain within 150 steps, so for P0.5 the flat greedy accuracy is not specific to the judge. Longer gold training may still help P0.5. At step 50, all gold cells also classify as sharpening, so the early T1 rise is a property of this short recipe, not a sign of judge exploitation. At step 50, the greedy gain of P0.5 under gold reward minus that under Q3 is $+0.016$ $[-0.035, 0.068]$ (200 problems).
In 33 of 42 runs, onset $\tthree$ came before the first step of the run's 10-step window with the highest rollout accuracy, so onset does not mark where the reward stops helping. Absolute FPR in the plateau window is $0.47$--$0.54$ under Q7 and $0.76$--$0.99$ under the other judges.

\begin{figure}[t]
\noindent\fbox{\parbox{0.97\linewidth}{\small\textbf{P1.5$\times$Q7, training step 57. Reference answer 41. Model answer 31. $G_{\text{cal}}=0.82$, accepted.}\\[2pt]
\textit{Question.} Randy drew 5 pictures. Peter drew 3 more pictures than Randy. Quincy drew 20 more pictures than Peter. How many pictures did they draw altogether?\\[2pt]
\textit{Answer.}\\
{\ttfamily\footnotesize If Randy drew 5 pictures, then Peter drew 5 + 3 = 8 pictures.\\
Quincy drew 8 + 20 = 28 pictures.\\
To find the total number of pictures drawn by all of them, we add up the number of pictures each person drew: 5 (Randy) + 8 (Peter) + 28 (Quincy) = 31 pictures.\\
\#\#\#\# 31}}}
\caption{\textbf{A fluent wrong answer that the strong judge accepts.} The 1.5B policy under the Qwen2.5-7B judge at training step 57; the last sum is wrong ($5+8+28=41$, not $31$).}
\label{fig:fooled}
\end{figure}

\textbf{What a fooled answer looks like.} The strong judge Q7 accepts this answer from P1.5 at training step 57 (Figure~\ref{fig:fooled}; Appendix~\ref{app:examples} shows more examples, the step-0 answers of the same policy, a random sample and an error-type count). Every step is correct except the last sum ($5+8+28=41$, not $31$).

\begin{table}[t]
\caption{\textbf{Gold accuracy per cell} (seed means). Held-out: greedy, 200 problems, lenient label. Max gain: largest gain over step 0 at any evaluation step (and its step). Acceptance: share of held-out answers the training judge accepts. Rollout: lenient accuracy of training rollouts, steps 1--10 and the highest 10-step window (window end). Acceptance was not logged in stage 6. Gold: gold-reward control (stage 7; 2 seeds, logged values; no judge reward, so no acceptance).}
\label{tab:gold}
\centering
\small
\setlength{\tabcolsep}{4pt}
\begin{tabular*}{\linewidth}{@{\extracolsep{\fill}}llrrrrr}
\toprule
Judge & Policy & Step 0 & Max gain (step) & Step 150 & Acceptance 0 $\to$ 150 & Rollout 1--10 $\to$ peak \\
\midrule
Q1.5 & 0.5B & 0.33 & $+0.015$ (20) & 0.09 & 0.68 $\to$ 1.00 & 0.22 $\to$ 0.31 (40) \\
Q1.5 & 1.5B & 0.61 & $+0.037$ (40) & 0.43 & 0.78 $\to$ 0.99 & 0.49 $\to$ 0.66 (50) \\
Q3 & 0.5B & 0.33 & $+0.000$ (30) & 0.25 & 0.57 $\to$ 0.91 & 0.24 $\to$ 0.39 (40) \\
Q3 & 1.5B & 0.61 & $+0.030$ (70) & 0.59 & 0.69 $\to$ 0.90 & 0.51 $\to$ 0.74 (90) \\
Q7 & 0.5B & 0.33 & $+0.026$ (110) & 0.33 & 0.47 $\to$ 0.58 & 0.23 $\to$ 0.47 (130) \\
Q7 & 1.5B & 0.61 & $+0.034$ (60) & 0.61 & 0.66 $\to$ 0.78 & 0.50 $\to$ 0.72 (110) \\
L3 & 0.5B & 0.33 & $-0.021$ (20) & 0.11 & 0.60 $\to$ 0.91 & 0.22 $\to$ 0.29 (110) \\
L3 & 1.5B & 0.61 & $-0.005$ (20) & 0.44 & 0.65 $\to$ 0.93 & 0.49 $\to$ 0.57 (40) \\
L8 & 0.5B & 0.33 & $+0.032$ (60) & 0.30 & 0.60 $\to$ 0.88 & 0.22 $\to$ 0.44 (120) \\
O1 & 0.5B & 0.33 & $-0.038$ (5) & 0.02 & 0.62 $\to$ 1.00 & 0.22 $\to$ 0.22 (10) \\
O7 & 0.5B & 0.33 & $-0.018$ (35) & 0.21 & 0.59 $\to$ 0.94 & 0.23 $\to$ 0.35 (40) \\
Qw3-0.6 & 1.5B & 0.61 & $+0.003$ (30) & 0.465 & -- & 0.50 $\to$ 0.63 (40) \\
\midrule
Gold & 0.5B & 0.33 & $+0.048$ (100) & 0.32 & -- & 0.23 $\to$ 0.47 (120) \\
Gold & 1.5B & 0.61 & $+0.055$ (120) & 0.65 & -- & 0.52 $\to$ 0.75 (150) \\
\bottomrule
\end{tabular*}
\end{table}

\section{\texorpdfstring{In our cells with judge $\geq$ policy, the gap did not set the pace}{In our cells with judge >= policy, the gap did not set the pace}}
\label{sec:gap}

\begin{table}[t]
\caption{\textbf{Step-0 judge statistics and hacking endpoints for all 12 cells.} Ratio = judge/policy parameter count. AUC and $\FPR_0$ at step 0. $\thalf$ per seed (``$>T$'' = censored). GM = geometric mean from the 38-run AFT fit ($^{*}$: 42-run fit, stage 6). $\Hbar$: seed mean (lenient label). FPR: absolute FPR in steps 101--150.}
\label{tab:cells}
\centering
\small
\setlength{\tabcolsep}{4pt}
\begin{tabular*}{\linewidth}{@{\extracolsep{\fill}}llrrrlrrrr}
\toprule
Judge & Policy & Ratio & AUC & $\FPR_0$ & $\thalf$ per seed & GM $\thalf$ & GM $\tthree$ & $\Hbar$ & FPR \\
\midrule
Q1.5 & 0.5B & 3.1 & 0.740 & 0.449 & 18, 21 & 19.4 & 14.0 & 0.975 & 0.99 \\
Q1.5 & 1.5B & 1.0 & 0.677 & 0.555 & 22, 20 & 21.0 & 17.0 & 0.965 & 0.98 \\
Q3 & 0.5B & 6.3 & 0.844 & 0.265 & 27, 23, 21, 34 & 25.8 & 17.4 & 0.816 & 0.86 \\
Q3 & 1.5B & 2.0 & 0.808 & 0.359 & 30, 41, 45, 34 & 37.0 & 22.5 & 0.654 & 0.78 \\
Q7 & 0.5B & 15.5 & 0.885 & 0.161 & $>$150, $>$150, $>$250, 119 & 192 & 49.4 & 0.374 & 0.47 \\
Q7 & 1.5B & 4.9 & 0.873 & 0.218 & 88, 90, 51, $>$150 & 89.9 & 44.8 & 0.414 & 0.54 \\
\midrule
L3 & 0.5B & 6.6 & 0.762 & 0.450 & 26, 44, 45, 45 & 39.0 & 18.7 & 0.834 & 0.91 \\
L3 & 1.5B & 2.1 & 0.697 & 0.539 & 46, 45 & 45.5 & 34.4 & 0.802 & 0.91 \\
\midrule
L8 & 0.5B & 16.4 & 0.842 & 0.282 & 56, 40, 48, 34 & 43.7 & 23.0 & 0.673 & 0.76 \\
O1 & 0.5B & 3.0 & 0.759 & 0.430 & 15, 23, 19, 20 & 19.0 & 12.9 & 0.989 & 0.99 \\
O7 & 0.5B & 14.9 & 0.829 & 0.268 & 26, 25, 22, 26 & 24.7 & 17.2 & 0.853 & 0.89 \\
\midrule
Qw3-0.6 & 1.5B & 0.39 & 0.702 & 0.547 & 29, 32, 26, 35 & 30.3$^{*}$ & 18.0$^{*}$ & 0.980 & 0.99 \\
\bottomrule
\end{tabular*}
\end{table}

\begin{figure}[t]
\centering
\includegraphics[width=\linewidth]{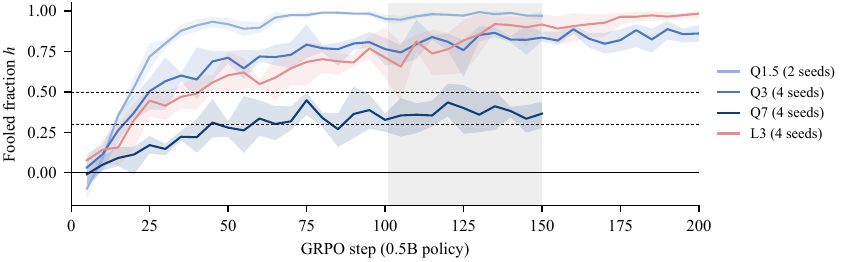}
\caption{\textbf{Fooled fraction $h$ for the 0.5B policy} under the four stage 1--3 judges (seed mean and range). Gray: plateau window. Dashed: the onset thresholds $0.3$ and $0.5$; $0.5$ lies just above the Q7 plateau. Q1.5 and Q7 curves end at step 150, the shortest run in those cells.}
\label{fig:curves}
\end{figure}

The depth of judge hacking (the plateau) tracks the judge: it correlates with judge AUC (Spearman $-0.69$ $[-0.94, -0.14]$ over 12 cells), and within a family the larger judge is fooled less deeply (Section~\ref{sec:scale}). For the pace (onset), the two pre-specified pairs contradict the gap account, but with 12 cells we cannot formally separate judge AUC from the gap for either endpoint (difference in correlations: plateau $-0.38$ $[-0.95, 0.00]$; onset $0.36$ $[0.00, 0.91]$).

\textbf{Pre-specified pairs.} P0.5$\times$Q3 has a \emph{larger} gap than P1.5$\times$Q7 in parameters (6.3 vs.\ 4.9) and in step-0 skill (judge AUC $0.844$ vs.\ $0.873$; policy accuracy $0.33$ vs.\ $0.61$). The gap account predicts the same or a later onset. It is fooled sooner and deeper on every endpoint (Table~\ref{tab:pairs}): $2.6\times$ sooner at $h\geq0.3$ ($3.5\times$ at the pre-specified $h\geq0.5$), $2.3\times$ in sampled KL ($5.0$--$7.3$ vs.\ $13.5$--$15.8$ nats; 2 of 4 P1.5$\times$Q7 runs have sampled KL), and $0.40$ higher in plateau. We lead with $h\geq0.3$ because $0.5$ sits just above the Q7 plateau (Figure~\ref{fig:curves}). Pair 2 gives the same sign. Both pairs were specified before stage 2 as \emph{descriptive} contrasts at $\thalf$; their $h\geq0.3$, KL and plateau versions were added after stage 5 (Appendix~\ref{app:prereg}).

\textbf{Over 12 cells, only the plateau--AUC correlation excludes zero.} Over all 12 cells (onsets from the 42-run fit; 95\% intervals from a bootstrap over cells, 10{,}000 resamples), the Spearman correlation of log-onset $\tthree$ is $0.15$ $[-0.52, 0.72]$ with the capability gap and $0.51$ $[-0.18, 0.92]$ with judge AUC alone; for plateau it is $-0.30$ $[-0.80, 0.34]$ and $-0.69$ $[-0.94, -0.14]$. Only the plateau correlation with AUC excludes 0 (Fisher $z$ intervals agree; Appendix~\ref{app:details}). The two onset correlations cannot be separated with 12 cells: their difference is $0.36$, paired-bootstrap interval $[0.00, 0.91]$. The gap is logit AUC minus a constant per policy, so the two correlations differ only through the policy main effect. Cells share judges and policies, so these intervals are optimistic. P0.5$\times$Q1.5 and P1.5$\times$Q7 have similar gaps ($1.76$, $1.49$) but reach $h\geq0.3$ at 14 and 45 steps (Figure~\ref{fig:main}; all cells in Table~\ref{tab:cells}). The judge's own solve accuracy was logged only in stage 6 (Q1.5: $0.605$; Qw3-0.6: $0.585$), so we could not use it as the judge skill.

\textbf{A judge smaller than the policy (stage 6).} In P1.5$\times$Qw3-0.6 the judge has $0.39$ times the policy's parameters and solves $0.585$ of the held-out problems (P1.5: $0.61$). Its step-0 AUC ($0.702$) is slightly above that of Q1.5 on the same policy ($0.677$; we computed no interval for the difference). The specification stated that this cell cannot decide between the two accounts, and its decision rule concerns only the AUC ordering. With this AUC the rule reverses the prediction: the judge account predicts a slower or shallower course than P1.5$\times$Q1.5, and a faster and deeper course would be evidence of a gap effect. The result matches neither pre-specified direction, and we report it as is. Onset at $h\geq0.3$ was not earlier (log-onset difference $+0.06$ $[-0.19, 0.31]$; 18.0 vs.\ 17.0 steps), and onset at $h\geq0.5$, the other pre-specified onset, was if anything later ($+0.37$ $[-0.00, 0.74]$, $p=0.051$; 30.3 vs.\ 21.0 steps). The plateaus were the same within error ($0.980$ vs.\ $0.965$; $+0.015$ $[-0.15, 0.18]$), but the control plateau leaves at most $0.035$ of room for a deeper course, so the onset comparison is the informative one. We see \emph{no evidence of a gap effect}. The cell also breaks the AUC ordering across families. Its AUC is close to that of P1.5$\times$L3 ($0.697$), but it was fooled sooner ($-0.65$ $[-0.89, -0.40]$ in log-onset) and deeper ($+0.18$ $[0.01, 0.34]$). The judge differs from the policy in family (Qwen3 vs.\ Qwen2.5) and in chat template (non-thinking), so this one cell cannot separate judge size from family. Fooled answers are again fluent wrong solutions, about as long as non-fooled ones (186 vs.\ 183 tokens), and we saw no token-level hacks, although the judge accepted 97\% of wrong answers cut off at 256 tokens (Appendix~\ref{app:extra}). The policy also drops ``\texttt{\#\#\#\#}'' after step 50 (rollout share $0.73$--$0.95$ in steps 1--50, $0$--$0.02$ in steps 51--100).

\textbf{Policy size under one judge.} The pre-registered primary rule returned R2, ``a policy effect only under the strong judge'' ($\thalf$: $-0.76$ $[-1.21, -0.36]$ at Q7). We do not adopt it: at the pre-registered threshold $h\geq0.3$ the Q7 contrast is $-0.10$ $[-0.33, 0.13]$ (97.5\%), in sampled KL $+0.09$ $[-0.46, 0.64]$, and the plateaus differ by $0.00$ under the parser-corrected label. At $h\geq0.3$ the pre-registered secondary contrast shows a policy effect under Q3 instead: $+0.26$ $[0.06, 0.46]$ (95\%; the larger policy is fooled later; Table~\ref{tab:contrasts}). Stage 1 had found that the policy effect depended on the coordinate (steps vs.\ greedy KL); its 8 runs lack sampled KL. The stage-3 Llama check is in Section~\ref{sec:scale}.

\section{A pre-registered blind test of step-0 predictors}
\label{sec:blind}

\begin{figure}[t]
\centering
\includegraphics[width=\linewidth]{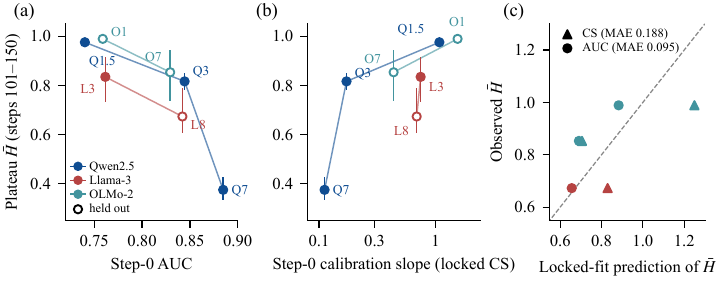}
\caption{\textbf{The locked step-0 predictor does not beat step-0 AUC on held-out judges.} P0.5 only. Filled: screening judges; open: held-out judges run blind. Lines join one family by size; bars: seed range. Plateau against (a) step-0 AUC and (b) the locked predictor CS (calibration slope, log scale). (c) Locked-fit predictions on the held-out cells (triangle: CS; circle: AUC; color: family; MAE: mean absolute error).}
\label{fig:blind}
\end{figure}

\textbf{Design.} On the 8 cells of Section~\ref{sec:gap} we screened 8 candidate step-0 predictors against 3 baselines (step-0 AUC, $\FPR_0$, and the risk score of \citet{zhou2026more}). A written rule selected CS, the calibration slope of step-0 correctness on $G$ ($\rho=0.95$ for plateau vs.\ $0.76$ for AUC), which mostly measures the judge's logit scale. CS and its 8-cell fit were locked by sha256. Separately, a hashed specification sent held-out judges that passed a step-0 AUC gate (${\geq}0.60$) to blind RL: L8, O1, O7 with P0.5, 4 seeds, 200 steps (Appendix~\ref{app:prereg}).

\textbf{Screening table.} Sign-adjusted 8-cell Spearman $\rho$ (plateau / onset) in stage 4: AUC 0.76/0.64, $\FPR_0$ 0.71/0.57, risk score 0.88/0.74; calibration slope (CS) 0.95/0.86; mass just below $\tau$ 0.81/0.64; non-correctness share of within-group score variance 0.95/0.83; reward standard deviation $-0.83/-0.71$ (opposite to the expected sign); wrong-last-step detection 0.57/0.36; format gain 0.67/0.48; surface gain 0.43/0.29; false positives after rewriting 0.60/0.45. Only CS and the format gain passed the leave-Llama-out check. The 8-cell linear fit of CS had a larger residual standard deviation (0.153) than the three baselines (0.113--0.119). Step-0 CS values (P0.5 / P1.5): Q1.5 1.078/0.851, L3 0.745/0.608, Q3 0.172/0.157, Q7 0.112/0.108. On the held-out cells, CS errs more than AUC on L8 ($0.157$ vs.\ $0.017$) and about the same on O7 ($0.148$ vs.\ $0.162$); over the 7 judges with P0.5 (descriptive), $\rho$ is $0.89$ for AUC and $0.82$ for P1.

\textbf{Result.} Held-out plateaus were $0.67$ (L8), $0.99$ (O1) and $0.85$ (O7) (Figure~\ref{fig:blind}). CS ordered 2 of 3 pairs ($\rho=0.5$), AUC all 3 ($\rho=1.0$). CS misordered the cross-family pair: O7 has the wider logit scale but is fooled more (O7 minus L8: $[0.037, 0.323]$). Plateau errors of the locked fits were $0.188$ (CS) and $0.095$ (AUC), partly from extrapolation; onset errors were equal. \textbf{Reading:} CS did not beat AUC on three held-out cells, and the test cannot separate them: $\rho\in\{\pm0.5,\pm1\}$, and a random predictor orders 3 cells perfectly with probability $1/6$. With one policy, it also could not test CS's one screening advantage (pooling the two policy rows).

\subsection{Judge families}
\label{sec:scale}

Within each family the larger judge was fooled less deeply (plateau difference, 95\%): Qwen 1.5B$-$3B $+0.16$ $[0.12, 0.20]$, 3B$-$7B $+0.44$ $[0.37, 0.51]$; Llama 3B$-$8B $+0.16$ $[0.02, 0.30]$; OLMo-2 1B$-$7B $+0.14$ $[0.00, 0.27]$ (the OLMo-2 interval touches 0). This agrees in direction with prior work in which larger reward models overoptimize less \citep{gao2023scaling} and a weakened frozen judge is hacked faster \citep{kenton2026debate}. Within Llama and OLMo-2, size is confounded with release. At $\thalf$ only Qwen 3B$\to$7B has a clear onset difference; at $\tthree$, OLMo-2 and Llama also differ (Table~\ref{tab:contrasts}). Across families, size does not set depth: at 7--8B the plateau is $0.37$ (Qwen), $0.67$ (Llama) and $0.85$ (OLMo-2).

\textbf{Second family (stage 3).} With L3 in place of Q3, the pre-registered P0.5 plateau is equivalent within $\pm0.15$ ($+0.018$ $[-0.106, 0.143]$), although L3's step-0 AUC ($0.762$) is close to Q1.5's ($0.740$) and L3 is less fooled than Q1.5 ($-0.14$ $[-0.27, -0.01]$). The equivalence holds only in that window: L3 is fooled more slowly early and keeps rising later ($h=0.947$ vs.\ $0.847$ for Q3 in steps 151--200, post hoc). Under L3, policies also move their final answer from ``\texttt{\#\#\#\# n}'' to ``\texttt{The answer is: n}'', a common Llama style.

\textbf{Offline transfer to Llama.} We re-scored fooled wrong answers from the Qwen-judge runs with Llama-3.2-3B-Instruct (no RL). Hacks found against Q1.5 transfer little: $h_L$ is $0.05$ (P0.5) and $-0.18$ (P1.5). Hacks found against Q3 and Q7 transfer in part ($h_L$ 0.07--0.36; the Q7 values include misread correct answers).

\begin{table}[t]
\caption{\textbf{Onset and plateau contrasts between cells} (natural-log scale for onsets; plateau in units of $h$). Onsets from the 38-run AFT fit. $\Delta_P$ = log-onset of the 1.5B policy minus the 0.5B policy under the same judge. Values from the registered stage-2 fit are in Appendix~\ref{app:prereg}.}
\label{tab:contrasts}
\centering
\small
\setlength{\tabcolsep}{4pt}
\begin{tabular*}{\linewidth}{@{\extracolsep{\fill}}lrll}
\toprule
Contrast & Estimate & Interval & Status \\
\midrule
Pair 1, $\thalf$: P0.5$\times$Q3 $-$ P1.5$\times$Q7 & $-1.25$ & $[-1.62, -0.88]$ (97.5\%) & descriptive, pre-specified \\
Pair 2, $\thalf$: P0.5$\times$Q1.5 $-$ P1.5$\times$Q3 & $-0.64$ & $[-1.09, -0.20]$ (97.5\%) & descriptive, pre-specified \\
$\Delta_P$ at Q1.5 & $+0.08$ & $[-0.44, 0.59]$ (97.5\%) & descriptive \\
$\Delta_P$ at Q3 & $+0.36$ & $[-0.00, 0.73]$ (97.5\%) & primary \\
$\Delta_P$ at Q7 & $-0.76$ & $[-1.21, -0.36]$ (97.5\%) & primary \\
$\Delta_P$ at Q7, $h\geq0.3$ & $-0.10$ & $[-0.30, 0.10]$ (95\%) & secondary, pre-registered \\
$\Delta_P$ at Q3, $h\geq0.3$ & $+0.26$ & $[0.06, 0.46]$ (95\%) & secondary, pre-registered \\
\midrule
Plateau, P0.5: L3 $-$ Q3 & $+0.018$ & $[-0.106, 0.143]$ (95\%) & primary (equivalence $\pm0.15$) \\
Plateau, P0.5: L3 $-$ Q1.5 & $-0.14$ & $[-0.27, -0.01]$ (95\%) & descriptive \\
Onset $\thalf$, P0.5: L3 $-$ Q3 & $+0.41$ & $[0.10, 0.73]$ (95\%) & secondary \\
Onset $\tthree$, P0.5: L3 $-$ Q3 & $+0.07$ & $[-0.13, 0.27]$ (95\%) & secondary \\
\midrule
Onset $\thalf$, P0.5: Q1.5 $-$ Q3 & $-0.28$ & $[-0.67, 0.10]$ (95\%) & descriptive \\
Onset $\thalf$, P0.5: L3 $-$ L8 & $-0.11$ & $[-0.43, 0.20]$ (95\%) & descriptive \\
Onset $\thalf$, P0.5: O1 $-$ O7 & $-0.26$ & $[-0.58, 0.06]$ (95\%) & descriptive \\
Onset $\tthree$, P0.5: Q1.5 $-$ Q3 & $-0.22$ & $[-0.47, 0.02]$ (95\%) & descriptive \\
Onset $\tthree$, P0.5: L3 $-$ L8 & $-0.21$ & $[-0.41, -0.01]$ (95\%) & descriptive \\
Onset $\tthree$, P0.5: O1 $-$ O7 & $-0.29$ & $[-0.49, -0.09]$ (95\%) & descriptive \\
\bottomrule
\end{tabular*}
\end{table}

\section{Related work and positioning}
\label{sec:related}

The closest work is \citet{zhou2026more}: Qwen3 policies on GSM8K trained by self-play against reference-free judgments raise the judge's pass rate while accuracy stays flat, and the errors transfer across judge families. We see the same with separate frozen judges and add a crossed design. On transfer, hacks found against Q3 and Q7 partly fool Llama-3.2-3B ($h_L$ 0.07--0.36), but hacks found against Q1.5 hardly do ($h_L$ $0.05$ and $-0.18$). The error types do not point to hacks that exploit a weak judge's arithmetic slips: such slips are not more common under weaker judges (Q1.5 10/40, Q3 12/40, Q7 9/40; Table~\ref{tab:errtypes}). \citet{kenton2026debate} study a judge weaker than the policy; our one cell with a smaller judge solves about as well as the policy and comes from another family, so it is not a clean test of that regime. \citet{huang2025accuracy} find that more accurate model-based math verifiers are still hacked in RL, especially after fine-tuning, so our AUC result applies only to zero-shot instruct judges. A static math reward benchmark tracks best-of-$n$ policy outcomes \citep{kim2024evaluating}, and \citet{zhang2026leaky} run a pre-registered GRPO study on code in which a pre-training audit tracks rewarded false positives; we differ in using LLM judges, judge-side step-0 metrics and a hash-locked selection before a blind test. More capable agents can exploit misspecified rewards more, with phase transitions \citep{pan2022effects}; two policy sizes cannot exclude such a threshold. \citet{anon2025genrm} varied policy and generative reward-model size (Qwen3 0.6B--14B, creative writing, online DPO) and found reward-model scale most decisive, even when the reward model far exceeds the policy; the submission was not accepted at ICLR 2026.

\textbf{Overoptimization and predicting RL outcomes.} \citet{gao2023scaling} showed that the coefficients of the overoptimization curve of a learned proxy reward model change smoothly with reward-model size, and that a 6B policy overoptimizes about as much as a 1.2B policy (two policy sizes; they note that this contradicts other internal findings). Our policy-size contrasts are mixed (an effect under Q7 at $h\geq0.5$ and under Q3 at $h\geq0.3$, Section~\ref{sec:gap}); we differ in the judge (a prompted, frozen LLM without a reference), the endpoint (false positives on wrong answers), the design (equal-gap cells), and the outcome (no gold improvement). Static benchmarks such as RewardBench \citep{lambert2024rewardbench} are standard; several works ask whether such metrics predict RL outcomes \citep{frick2024evaluate,kim2024evaluating,wen2025rethinking,kim2025rethinking,razin2025what}. Our blind test is a small pre-registered instance for frozen LLM judges.

\textbf{Hacking LLM judges and oversight.} LLM judges have known biases \citep{zheng2023judging} and can be fooled by short strings \citep{zhao2025one}. AdvJudge-Zero \citep{li2025advjudge} finds short control tokens that flip judge verdicts and shows that a hardened judge removes false-positive spikes under GRPO on MATH and GSM8K; its hacks are token-level, while ours are fluent, confident solutions with a wrong setup or an arithmetic slip (Appendix~\ref{app:extra}). RL can make answers more convincing without making them more correct \citep{wen2024mislead,zhou2026more}. \citet{liu2026examining} find that non-reasoning judges are hacked easily in non-verifiable post-training, while reasoning judges lead to policies that do well under a gold judge partly through adversarial outputs. \citet{kenton2026debate} train against a frozen judge weaker than the policy and find that weakening the judge further speeds up hacking, which agrees in direction with our within-family trend. A gap-based model of oversight \citep{engels2025scaling} motivates the gap account; weak-to-strong work studies what a weaker supervisor can elicit from a stronger model \citep{burns2023weak}.

\textbf{Other related work.} Scalable-oversight protocols \citep{bowman2022measuring,khan2024debating,kenton2024scalable} study how a weaker supervisor can oversee a stronger model, and \citet{engels2025scaling} model oversight success as a function of the capability gap in task-specific Elo. Building on the setup of \citet{gao2023scaling}, reward-model ensembles were proposed as a mitigation \citep{coste2024reward}, and later work studied overoptimization in direct alignment methods \citep{rafailov2024scaling}. LLMs can also be trained as verifiers or reward models \citep{cobbe2021training,lightman2023lets}, including generative verifiers \citep{zhang2024generative}. Our judges are untrained instruct models used zero-shot; training them, or letting them commit to their own answer first \citep{zhou2026more}, are defenses we do not test.

\section{Limitations and conclusion}
\label{sec:limits}

\textbf{Limitations.} (i) One task, prompt and recipe; no KL penalty; single-token judges without chain of thought; policies ${\leq}1.5$B, judges ${\leq}8$B. Only one judge is smaller than the policy (0.6B vs.\ 1.5B); it solves about as well ($0.585$ vs.\ $0.61$) and comes from another family and template, so a judge weaker than the policy is tested only in one confounded cell. (ii) 2--4 seeds per cell; the predictor test has 3 held-out cells and one policy. (iii) Screening candidates, the plateau endpoint, the test's pass rule and the step coordinate were set after some outcomes were known (Appendix~\ref{app:prereg}). (iv) Gold labels are parsers (5.7\% errors in the per-answer check, done by an LLM agent and by the author). (v) The gold-reward control has 2 seeds, 150 steps and one learning rate; the 12-cell correlation intervals treat cells as independent.

\textbf{Conclusion.} In our 12 cells of GRPO against frozen, reference-free LLM judges, judge-RL did not detectably improve held-out greedy accuracy. A gold-reward control with the same recipe did improve it for P1.5, so there the flat accuracy is specific to the judge reward; for P0.5 it did not, so there it is not. In the four rising cells we checked, the early rise in training-rollout accuracy was mostly sampling sharpening, as it was under gold reward. Where the judge was at least as large as the policy, the depth of judge hacking (the plateau) tracks the judge: it correlates with judge AUC (Spearman $-0.69$ $[-0.94, -0.14]$ over 12 cells), and within a family the larger judge is fooled less deeply. For the pace (onset), the two pre-specified pairs contradict the gap account, but with 12 cells we cannot formally separate judge AUC from the gap for either endpoint (difference in correlations: plateau $-0.38$ $[-0.95, 0.00]$; onset $0.36$ $[0.00, 0.91]$); this rests on descriptive pairs and 2--4 seeds per cell. One cell with a smaller judge gave no evidence of a gap effect, but its control plateau was near the ceiling and it is confounded with judge family.

\raggedbottom
\section*{Acknowledgments}

Claude Opus 5.5 (Anthropic), working through Claude Code, performed all computations and manuscript preparation. H.W. conceived and directed the project, checked all results and the manuscript, and takes full responsibility for the content of this paper.

\section*{Author contributions}

H.W. conceived and directed the project. He proposed the research questions, the core ideas and the hypotheses, designed each stage, reviewed the results of every stage, and set the overall direction. H.W. and Claude interpreted the results jointly. Claude Opus 5.5, an AI model developed by Anthropic, used through Claude Code, performed all computations: it wrote and debugged all code, launched and monitored all runs, computed all statistics, and did the per-answer reading. Claude also wrote the stage specifications from H.W.'s designs, surveyed the literature, checked every citation, made all figures and tables, and wrote the manuscript. H.W. shaped the arguments, the text and the figures through repeated discussion with Claude, and Claude revised them after each round. H.W. checked all figures and numbers, read all answer samples, read the whole manuscript, and is fully responsible for the scientific content and integrity of this paper.

\section*{AI use disclosure}

The author conceived and directed the project and made all scientific decisions. LLM agents were used extensively under the author's continuous direction. The main agent was Claude Opus 5.5 (Anthropic), run through Claude Code. It started sub-agents (other Claude instances, including Claude Sonnet for literature search and classification) for parallel tasks, and several agent sessions worked in parallel on different stages, coordinated through shared status files.

The human author conceived and directed the project. The author proposed the research questions, the core ideas and the hypotheses, designed each stage, set the overall direction, reviewed the results of every stage and decided whether to continue. The author and the agents interpreted the results jointly. The agents did the following tasks:
\begin{itemize}\setlength{\itemsep}{0pt}
\item \textbf{Specifications:} wrote the stage specifications from the author's designs. The author reviewed each one, and each was locked with a sha256 hash before its runs (Appendix~\ref{app:prereg}).
\item \textbf{Implementation:} wrote and debugged all training, scoring and analysis code, and launched and monitored all runs on the GPU servers.
\item \textbf{Analysis:} computed all statistics and did the per-answer reading (the 300-answer parser audit and the 20 fooled answers per cell, Appendix~\ref{app:extra}).
\item \textbf{Literature:} surveyed the literature and checked every citation against its arXiv, proceedings or model-card page.
\item \textbf{Writing and figures:} drafted and polished the text, figures and tables under the author's direction. The author shaped the arguments and reviewed every draft, and the agents revised the drafts after each round of feedback. Simulated reviewer reports, also generated by agents, were used in the revision.
\end{itemize}
\textbf{Verification.} An agent traced every number in the paper to a saved result file, and a separate agent pass checked numbers, citations and claims before submission. The author checked all figures and numbers and read all answer samples. The author takes responsibility for the final content of this work, including text, claims and artifacts produced with the aid of generative AI. No AI system is listed as an author.

\section*{Ethics statement}

This study has no human subjects and uses no personal data: all answers were generated by open-weight models on GSM8K problems. The licenses of all models and data are listed in Appendix~\ref{app:details}. Our results show that zero-shot LLM judges for math are easy to exploit with RL, and we report this as a caution for those who use such judges as rewards. The hacks we observed are ordinary wrong solutions, and we introduce no new attack method.

\section*{Reproducibility statement}

Section~\ref{sec:setup} and Appendix~\ref{app:details} give the models (with revisions), data split, judge prompt, calibration, hyperparameters and compute. Appendix~\ref{app:prereg} lists every stage specification and deviation. Code, per-step logs and analysis scripts will be released with the paper.

\flushbottom
\addtolength{\bibsep}{0pt plus 1.5pt}
\bibliography{refs}
\bibliographystyle{iclr2027_conference}

\appendix

\section{Pre-registration record and deviations}
\label{app:prereg}

The study has seven stages. Each stage had a written specification with a recorded sha256 before its runs. Changes after that were written as dated addenda.

\textbf{Stage 1 (8 runs, Q1.5 and Q7).} The original endpoint used a \emph{strict} gold label (requires ``\texttt{\#\#\#\#}''). Its pre-registered verdict was ``depends on the coordinate'': in steps the judge account won, in (greedy) KL both factors contributed, and the Q1.5/Q7 contrast was about $3$--$4.5\times$ in steps but about $2.5\times$ in KL. A post-hoc check found that some policies dropped the answer format, so the strict label counted correct answers as wrong. This motivated the lenient label. All 8 runs were re-scored from saved rollouts.

\textbf{Stage 2 (12 runs, adds Q3 and seeds for Q7).} Written before any Q3 run. Fixed in advance: the lenient label, steps as the coordinate (chosen after stage 1), the AFT model, the 97.5\% level, decision rules R1--R4, the two equal-gap pairs as descriptive contrasts, and a 24 A100-hour cap. Deviations: (1) the earliest design planned a MATH subset; stage 1 switched to GSM8K, registered before stage 2. (2) The prior $\ln\sigma\sim\mathcal{N}(\ln0.25,0.5^2)$ was set after the lenient re-scoring of the 8 old runs, before any Q3 data; without it, $\sigma=0.274$ and results are almost identical. (3) A \emph{generous} gold label (accepts a correct number in the last sentence) was added before any Q3 result; the verdict did not change. (4) After 2 runs, ties at $\tau$ (differences near $10^{-8}$) were resolved with $G_{\text{cal}}+10^{-6}\geq\tau$; no onset changed. (5) A planned third seed for the Q1.5 row was not run because of the compute cap. (6) Old P0.5$\times$Q7 seeds stop at 150 steps; new seeds ran to 250. Verdict: R2 by the written rule. The registered stage-2 fit (20 runs, $\sigma=0.27$) gave: equal-gap pairs (larger-gap cell first) $-1.26$ $[-1.73, -0.80]$ and $-0.64$ $[-1.21, -0.08]$; $\Delta_P$ at Q1.5, Q3, Q7: $+0.08$, $+0.36$ $[-0.10, 0.82]$, $-0.79$ $[-1.38, -0.30]$ (97.5\%). This paper reports all stage 1--5 onsets from one AFT fit over those 38 runs (same model and prior, $\sigma=0.222$); the signs and verdicts are the same.

\textbf{Stage 3 (6 runs, L3).} Written after the static gate and two smoke runs, before any analyzed run. The plateau endpoint and window were chosen \emph{after} stage 2 showed the Q7 plateau, but before any Llama data. The Qwen control values were known when the margin $\pm0.15$ was set. Verdict: V1 (``by size''); the upper bound ($0.143$) is $0.007$ from the margin.

\textbf{Stage 4 (screening, no RL).} Written before the scoring jobs. The candidate list, outcomes (plateau and log-onset at $h\geq0.3$), selection rule, tie rule, leave-Llama-out check and stopping rules were fixed in the specification. The outcomes and the earlier findings were known when the candidates were defined; the format-gain candidate in particular came from the Llama format drift. The selected predictor and its fits (outcome $=\alpha+\beta x+\gamma[\text{P1.5}]$) were written to a file whose sha256 (\texttt{8c1eb85a\ldots}) was recorded before unblinding.

\textbf{Stage 5 (blind test, 12 runs).} The specification (judges, gate, recipe, endpoints, blinding rules) was hashed before any of its RL runs. All outputs went to a blind directory, and analysis started only after both hashes were re-checked. Deviations: (1) Llama-3.2-1B failed the gate and was not run. (2) The test protocol did not state its own pass rule; at unblinding we reused the stage-4 qualification rule (both $\rho>0$ and plateau error below the best baseline). CS fails it. (3) A repair agent once wrote a diagnostic run (not a study run) into the blind directory; it was deleted unread before unblinding.

\textbf{Stage 6 (4 runs and a check without training; added in revision).} The specification (sha256 \texttt{\small 1161ccdf0afc54373a1f65472dd2fb5a2a5f69007d1471e1aad5cafe4daa1477}) was written before the gate and before any stage-6 run. It fixed the gate (step-0 AUC ${\geq}0.60$ on P1.5; candidates in order Qwen2.5-0.5B-Instruct, then Qwen3-0.6B), the recipe, the endpoints, the prediction rule and the sharpening rule. Qwen2.5-0.5B-Instruct failed the gate (AUC $0.581$), so Qwen3-0.6B was used (AUC $0.702$). The specification stated that this cell cannot decide between the two accounts and gave a decision rule only for the AUC ordering. For a judge with a higher AUC than Q1.5, the rule reverses the prediction: the judge account then predicts a slower or shallower course than P1.5$\times$Q1.5, and a faster and deeper course would be evidence of a gap effect. The result matches neither pre-specified direction (onset not earlier at $h\geq0.3$, later at $h\geq0.5$ with $p=0.051$; plateau the same within error), and we report it as is. Sharpening check: both main cells are ``sharpening''. Deviations: (1) the SFT validity check of the sharpening check failed (Appendix~\ref{app:extra}); (2) the plateau comparison used the point estimate of P1.5$\times$Q1.5 ($0.965$, 2 seeds), not its interval; (3) refitting the shared-$\sigma$ AFT model with the 4 new runs moved the other cells slightly (largest change: P0.5$\times$Q7, $\thalf$ geometric mean $192.3\to191.0$); this paper keeps the 38-run fit for stages 1--5. Disclosed in the specification: one smoke run of the sharpening check (one non-main checkpoint) ran before it was written, and the stage-6 runs used one GPU instead of two (same mathematics).

\textbf{Stage 7 (4 runs and a check without training; added before submission).} The specification (sha256 \texttt{\small 91a7f9314aaaa92ab06698cacef09893fc124475a16822acf561214a1e9cce14}) was written before any stage-7 run. It fixed the gold reward (1 if the lenient label is correct, else 0; the judge is still loaded and its scores are only logged), the recipe (as in the main design), 2 seeds and 150 steps per policy, the 500-problem evaluation, the rule for ``detected learning'' (lower bound of the step-150 greedy gain above 0 and both seeds above 0), the paper wording for each outcome, and the sharpening check of P0.5$\times$Q7 at step 50 with the stage-6 rule. Outcome: learning detected for P1.5, not for P0.5, so the specification's ``split by policy'' wording applies. Its predictions were P1.5 not detected (wrong), P0.5 either way, and P0.5$\times$Q7 sharpening (right). Validity: the SFT P0.5 measurement matched stage 6 on all 200 shared problems, and all 4 P0.5$\times$Q7 step-50 checkpoints matched their logged greedy accuracy within $0.02$. Deviations: (1) the specification's paper wording said ``three'' checked rising cells; with P0.5$\times$Q7 there are four; (2) not pre-registered: the SFT strict greedy accuracy again differs from the logged step-0 value ($0.31$ and $0.58$ vs.\ $0.33$ and $0.61$, as in stage 6), and the gold runs' re-measured greedy accuracy differs from the logged value by at most $0.035$.

\textbf{Post-hoc analyses (no new runs).} The following were added after the stage-5 analysis: the gold-accuracy analysis (Section~\ref{sec:gold}), the onset-versus-rollout-peak count, the capability gap, the $\tthree$, KL and plateau versions of the two pairs, the parser audit, and the 12-cell correlations. They use only saved logs and saved generations.

\section{Experimental details}
\label{app:details}

\begin{table}[ht]
\caption{\textbf{Hyperparameters} (from the saved run configurations). Both policies use the same SFT recipe. All 42 RL runs share the GRPO values (the stage-6 runs used one GPU instead of two). ``Default'': library default of TRL 1.14.1 and transformers 5.18.0, which all runs used.}
\label{tab:hparams}
\centering
\small
\setlength{\tabcolsep}{4pt}
\begin{tabular*}{\linewidth}{@{\extracolsep{\fill}}lll}
\toprule
 & SFT & GRPO \\
\midrule
Initial model & Qwen2.5-0.5B / 1.5B (base) & the SFT checkpoint \\
Data & 2{,}000 GSM8K solutions & 5{,}473 GSM8K problems \\
Length & 1 epoch & 150, 200 or 250 steps \\
Batch & 16 sequences (8 $\times$ 2 accum.) & 64 rollouts $=$ 8 problems $\times$ 8 (2 $\times$ 32 accum.) \\
Learning rate & $1\times10^{-5}$ & $3\times10^{-6}$ \\
Schedule & cosine, 6 warm-up steps & constant, 3 warm-up steps \\
Optimizer & AdamW (default) & AdamW (fused implementation) \\
Loss & cross-entropy on target tokens & DAPO token-level loss \\
Advantage & -- & reward minus group mean, divided by group std. \\
KL penalty & -- & none ($\beta=0$) \\
Reward & -- & $G_{\text{cal}}$ of the frozen judge; no format reward \\
Maximum length & 512 tokens (prompt and target) & 256 new tokens \\
Sampling & -- & temperature 1.0, no top-$k$ or top-$p$ \\
Precision & fp32 & policy fp32, judge bf16 \\
Seeds & seed 2 & 301--304 \\
\bottomrule
\end{tabular*}
\end{table}

\begin{table}[ht]
\caption{\textbf{Notation.}}
\label{tab:notation}
\centering
\small
\begin{tabularx}{\linewidth}{@{}>{\raggedright\arraybackslash}p{0.16\linewidth}>{\raggedright\arraybackslash}X@{\hspace{1.5em}}>{\raggedright\arraybackslash}p{0.11\linewidth}>{\raggedright\arraybackslash}X@{}}
\toprule
Symbol & Meaning & Symbol & Meaning \\
\midrule
P0.5, P1.5 & policies: Qwen2.5-0.5B and -1.5B after SFT & $h$ & fooled fraction $(\FPR-\FPR_0)/(1-\FPR_0)$ on 5-step windows \\
Q1.5, Q3, Q7, L3, L8, O1, O7 & judges: family (Qwen2.5, Llama, OLMo-2) and size in billions & $\thalf$, $\tthree$ & onset: first step with $h\geq0.5$ ($0.3$) on two consecutive windows \\
Qw3-0.6 & judge: Qwen3-0.6B, non-thinking chat template (stage 6) & $\Hbar$ & plateau: $h$ over steps 101--150 \\
$G$ & judge score $\logit(\texttt{Yes})-\logit(\texttt{No})$ & AUC & step-0 judge AUC for correct versus wrong answers of the policy \\
$G_{\text{cal}}$ & calibrated score $\sigma(aG+b)$; the RL reward & Capability gap & $\logit(\text{AUC})-\logit(\text{step-0 greedy accuracy})$ \\
$\tau$ & per-cell acceptance threshold (true-positive rate $0.80$ at step 0) & CS & calibration slope of step-0 correctness on $G$ \\
$\FPR$, $\FPR_0$ & share of wrong answers with $G_{\text{cal}}\geq\tau$; its step-0 value & $\Delta_P$ & log-onset of P1.5 minus log-onset of P0.5 under the same judge \\
\bottomrule
\end{tabularx}
\end{table}

\textbf{Policy prompt and SFT data.} The policy sees plain text without a chat template, in SFT, in RL and in evaluation:

\noindent\fbox{\parbox{0.97\linewidth}{\ttfamily\footnotesize Solve the math problem. Show your reasoning, then give the final answer on the last line in the form "\#\#\#\# <number>".\\[4pt]
Problem: \{q\}\\
Solution:}}
\vspace{4pt}

The SFT target is a space followed by the GSM8K reference solution with its calculator annotations (\texttt{<<\ldots>>}) removed; it keeps the native last line ``\texttt{\#\#\#\# n}''. The loss is on the target tokens only. The first of the 2{,}000 SFT examples (target only; the prompt is the template above with this problem):

\noindent\fbox{\parbox{0.97\linewidth}{\small\textit{Problem.} The result from the 40-item Statistics exam Marion and Ella took already came out. Ella got 4 incorrect answers while Marion got 6 more than half the score of Ella. What is Marion's score?\\[2pt]
\textit{Target.}\\
{\ttfamily\footnotesize Ella's score is 40 items - 4 items = 36 items.\\
Half of Ella's score is 36 items / 2 = 18 items.\\
So, Marion's score is 18 items + 6 items = 24 items.\\
\#\#\#\# 24}}}
\vspace{4pt}

\textbf{Statistical analysis.} \emph{Onsets.} For run $i$, let $y_i=\ln t_i$, where $t_i$ is its onset. A run with no onset by step $T_i-5$ is right-censored at $y_i=\ln T_i$ ($T_i$: its last step). The model has one mean $\mu_c$ per cell and one common $\sigma$. With $z_i=(y_i-\mu_{c(i)})/\sigma$, the log-likelihood is
\begin{equation*}
\ell(\mu,\sigma)=\sum_{i\ \text{not censored}}\bigl[\ln\phi(z_i)-\ln\sigma\bigr]+\sum_{i\ \text{censored}}\ln\bigl[1-\Phi(z_i)\bigr],
\end{equation*}
where $\phi$ and $\Phi$ are the standard normal density and distribution function. We maximize $\ell-(\ln\sigma-\ln0.25)^2/(2\cdot0.5^2)$: this is the prior $\ln\sigma\sim\mathcal{N}(\ln0.25,0.5^2)$ with flat priors on the $\mu_c$ (L-BFGS-B, then Nelder--Mead, from two starting values of $\sigma$). For a contrast $\delta=\mu_a-\mu_b$, the profile-likelihood interval is the set of $\delta$ with $2[\hat\ell-\hat\ell(\delta)]\leq\chi^2_{1,q}$, where $\hat\ell(\delta)$ is the maximum under the constraint $\mu_a-\mu_b=\delta$, and $q$ is $0.95$ or $0.975$ (Bonferroni over the two primary contrasts). A bound is reported as infinite if the profile does not reach the cut-off within $\pm9$ of the estimate. The geometric mean of a cell is $\exp(\hat\mu_c)$. \emph{Plateaus.} The unit is the seed: each run gives one $\Hbar$, and differences use Welch $t$ intervals with Welch--Satterthwaite degrees of freedom. \emph{KL at onset.} The KL of each run is linearly interpolated between evaluation steps to its onset step; differences use Welch intervals on the logarithm. \emph{Correlations} over the 12 cell means (onsets from the 42-run fit) are Spearman coefficients with 95\% percentile intervals from a bootstrap over cells (10{,}000 resamples). Fisher $z$ intervals (standard error $1.03/\sqrt{n-3}$) agree: onset $\tthree$ with the gap $[-0.48, 0.68]$ and with AUC $[-0.11, 0.84]$; plateau with the gap $[-0.75, 0.35]$ and with AUC $[-0.91, -0.17]$. At $\thalf$ the onset correlations are $0.17$ $[-0.49, 0.74]$ (gap) and $0.50$ $[-0.20, 0.91]$ (AUC), and their difference is $0.33$ $[-0.01, 0.86]$ (paired bootstrap); for plateau the difference is $-0.39$ $[-0.95, 0.00]$. Both methods treat cells as independent, but cells share judges and policies, so the intervals are too narrow. \emph{Step-0 predictors} use a problem-clustered bootstrap.

\textbf{Hardware and compute.} NVIDIA A100-PCIE-40GB GPUs. A100-hours = wall-clock hours $\times$ GPUs held, including failed launches. Stage 1: 13.45; stage 2: 22.9; stage 3: 8.55; stage 4: 0.44; stage 5: about 11.5; stage 6: about 5.3 (RL 4.22, sharpening check 0.83, gate and smoke runs about 0.25); stage 7: about 5.3 (gold-reward runs 0.89 and 0.93 for P0.5, 2.72 for P1.5; P0.5$\times$Q7 sharpening check 0.76). Total: about 67 A100-hours. Apart from stages 6 and 7, the revision used no GPU.

\textbf{Assets.} Qwen2.5-0.5B/1.5B (base), Qwen2.5-0.5B/1.5B/7B-Instruct, Qwen3-0.6B (revision c1899de2) and OLMo-2 1B/7B-Instruct: Apache-2.0. Qwen2.5-3B-Instruct (revision aa8e7253): Qwen Research License (research use only). Llama-3.2-3B-Instruct and Llama-3.1-8B-Instruct: Llama 3.2 and 3.1 Community Licenses. GSM8K (revision 740312ad): MIT. Code and per-step logs will be released with the paper.

\section{Additional results}
\label{app:extra}

\textbf{Parser audit (revision).} For each cell whose answer format drifts (P0.5 and P1.5 under Q7 and L3, P0.5 under L8 and O7), we drew 50 training rollouts at random from steps 101--150 that the lenient label calls wrong and the judge accepts, and read each one (an LLM agent read them, and the author also read all 300; see the AI use disclosure). An error is a solution whose stated final answer equals the reference. Errors: Q7 2/50 and 12/50 (P0.5, P1.5); L3 0/50 and 1/50; L8 0/50; O7 2/50; in total 17/300 (5.7\%). Every error has the same form: the final answer appears in prose and a later number follows it (``\ldots{} 240 oysters in 2 hours''). The generous label agreed with the per-answer label on all 300. Under the generous label, the P1.5$\times$Q7 plateau is $0.367$ instead of $0.414$; all other plateaus change by at most $0.007$. This correction makes the Q7 policy contrast $0.00$ and widens pair 1.

\textbf{Sampling-sharpening check (stage 6, no training).} For each checkpoint we decoded every held-out problem greedily and drew 4 samples at temperature 1. Each difference is the step-50 checkpoint minus the SFT policy, both measured in the same pass, with the lenient label and a problem-level paired bootstrap (95\%). The pre-registered rule calls a cell \emph{sharpening} if the lower bound of the temperature-1 gain is above 0 and the upper bound of the greedy gain is below the temperature-1 gain; it calls a cell \emph{learning} if the lower bound of the greedy gain is above 0. Table~\ref{tab:sharp} gives the result. Both main cells (Q3) are sharpening, and so are P0.5$\times$O7 and P0.5$\times$Q7 (stage 7; on 500 problems: $\Delta$T1 $+0.158$ $[0.136, 0.180]$, $\Delta$greedy $+0.026$ $[-0.009, 0.060]$, also sharpening). At step 150 (descriptive, 200 problems), the P0.5$\times$Q7 greedy gains per seed are $+0.065$, $+0.010$, $-0.015$ and $+0.045$; seeds 302 and 304 lost the answer format under the strict label (strict greedy accuracy $0.01$ and $0.06$), while their lenient greedy accuracy did not fall. Its rollout accuracy peaks near step 130 (seed mean). Under O1, where rollout accuracy fell, temperature-1 accuracy fell too. The per-token negative log-likelihood of the samples under the sampling policy (an entropy estimate) fell in all four cells, as sharpening predicts. pass@4 rose ($+0.10$ and $+0.20$ under Q3), against the specification's expectation. Most of the temperature-1 gain is on problems that greedy decoding already solves (P0.5: $0.33\to0.67$; P1.5: $0.49\to0.79$). SFT samples often miss such problems, so sharpening alone raises pass@4 at $k=4$. On problems that greedy decoding gets wrong, P1.5$\times$Q3 gained $+0.10$ ($0.116\to0.216$), a small improvement that does not change greedy accuracy. \emph{Validity.} The SFT greedy accuracy measured here ($0.31$ and $0.58$) differs from the logged step-0 value ($0.33$ and $0.61$) by more than the pre-registered $0.01$, so this check failed; we did not find the cause. The conclusion uses only differences whose two ends were measured on the same decoding path, and with the logged step-0 values the classification is unchanged. 15 of 16 step-50 checkpoints matched their logged greedy accuracy within $0.02$ (rule: at least 14). In the stage-6 RL runs, held-out temperature-1 accuracy (first 100 problems) rose from $0.37$ to $0.48$--$0.52$ (steps 35--90), while greedy accuracy on the same problems went from $0.61$ to $0.64$ (step 30) and then to $0.475$ (step 150).

\begin{table}[ht]
\caption{\textbf{Sampling-sharpening check} (stages 6 and 7). Upper block: step-50 checkpoint minus the SFT policy on the 200 held-out problems, lenient label, seed mean of 4 runs; 95\% problem-level paired bootstrap. T1: accuracy of 4 samples per problem at temperature 1. Greedy: greedy accuracy in the same pass. NLL: per-token negative log-likelihood of the samples (nats). Both: meets the rule for learning and for sharpening. Rollout: training-rollout accuracy, steps 41--50 minus steps 1--10. All pass@4 intervals exclude 0. SFT T1 / greedy accuracy: $0.147$ / $0.310$ (P0.5), $0.335$ / $0.580$ (P1.5). P0.5$\times$Q7: stage 7, same SFT measurement. Lower block: gold-reward control (stage 7), step-50 (c50) and step-150 (c150) checkpoints, 500 held-out problems (the first 200 are the ones above), 2 seeds, SFT re-measured in the same pass (T1 / greedy $0.138$ / $0.318$ for P0.5, $0.326$ / $0.574$ for P1.5); rollout rise not given for c150.}
\label{tab:sharp}
\centering
\small
\setlength{\tabcolsep}{2pt}
\begin{tabular}{lrrrrrl}
\toprule
Cell & $\Delta$T1 & $\Delta$greedy & $\Delta$pass@4 & $\Delta$NLL & Rollout & Class \\
\midrule
P0.5$\times$Q3 & $+0.129$ {\scriptsize$[0.094, 0.165]$} & $+0.004$ {\scriptsize$[-0.054, 0.062]$} & $+0.104$ & $-0.74$ & $+0.139$ & sharpening \\
P1.5$\times$Q3 & $+0.232$ {\scriptsize$[0.194, 0.268]$} & $+0.035$ {\scriptsize$[-0.024, 0.094]$} & $+0.195$ & $-0.51$ & $+0.228$ & sharpening \\
P0.5$\times$O7 & $+0.121$ {\scriptsize$[0.086, 0.156]$} & $-0.019$ {\scriptsize$[-0.076, 0.039]$} & $+0.071$ & $-0.78$ & $+0.091$ & sharpening \\
P0.5$\times$O1 & $-0.112$ {\scriptsize$[-0.146, -0.080]$} & $-0.276$ {\scriptsize$[-0.341, -0.212]$} & $-0.271$ & $-0.70$ & $-0.150$ & no rise \\
P0.5$\times$Q7 & $+0.147$ {\scriptsize$[0.113, 0.181]$} & $+0.026$ {\scriptsize$[-0.023, 0.075]$} & $+0.133$ & $-0.67$ & $+0.173$ & sharpening \\
\midrule
P0.5$\times$gold c50 & $+0.139$ {\scriptsize$[0.118, 0.161]$} & $+0.009$ {\scriptsize$[-0.029, 0.047]$} & $+0.153$ & $-0.61$ & $+0.133$ & sharpening \\
P0.5$\times$gold c150 & $+0.173$ {\scriptsize$[0.148, 0.198]$} & $+0.022$ {\scriptsize$[-0.015, 0.059]$} & $+0.157$ & $-0.77$ & -- & sharpening \\
P1.5$\times$gold c50 & $+0.198$ {\scriptsize$[0.175, 0.221]$} & $+0.026$ {\scriptsize$[-0.010, 0.063]$} & $+0.152$ & $-0.33$ & $+0.139$ & sharpening \\
P1.5$\times$gold c150 & $+0.278$ {\scriptsize$[0.253, 0.305]$} & $+0.054$ {\scriptsize$[0.013, 0.096]$} & $+0.175$ & $-0.52$ & -- & both \\
\bottomrule
\end{tabular}
\end{table}

\textbf{Common wrong problems.} $h$ on the 324 problems that both policies got wrong in every step-0 sample (0.5B / 1.5B policy), steps 1--50, 51--100, 101--150. Q1.5: 0.66/0.58, 0.94/0.92, 0.98/0.98. Q3: 0.49/0.35, 0.70/0.56, 0.83/0.66. Q7: 0.12/0.13, 0.32/0.40, 0.37/0.34. These are close to $h$ on all problems, and they show no advantage of P1.5 at Q7.

\textbf{What the hacks look like.} Appendix~\ref{app:examples} shows full examples. We read 20 fooled wrong answers per cell (pre-registered; read by an LLM agent and by the author). Under Q1.5, Q3 and L3, all 120 are wrong by the reference (one, under Q3, follows a plain reading of an ambiguous problem), and almost all keep ``\texttt{\#\#\#\#}''. They are fluent, confident solutions. Table~\ref{tab:errtypes} codes the first error of every answer: over all 220, it is most often a wrong setup (127; a missing step or a wrong operation), then an arithmetic slip (54; for example ``$3\times2000-500=550$'') and a misread quantity (28). There is \emph{no length hack}: fooled wrong answers are shorter than non-fooled ones in every cell of stages 1--5 (for example 124 vs.\ 146 tokens under Q3); in the stage-6 cell they are about as long (186 vs.\ 183 tokens). Under Q7, 12 of 20 (P0.5) and 13 of 20 (P1.5) fooled answers keep ``\texttt{\#\#\#\#}''. The generous label counts 2 and 4 of 20 as correct; on reading, 1 and 4 are correct answers that the lenient parser misreads, and under P1.5$\times$Q7 three more follow the problem but not the GSM8K reference. In the held-out cells, all 60 read answers are wrong, and the first error is most often a wrong setup (37 of 60). The 20 fooled answers of the stage-6 cell (not coded in Table~\ref{tab:errtypes}) are also fluent wrong solutions with a wrong setup or an arithmetic slip; in the onset windows, 32\% of its fooled wrong answers reach 250 or more tokens (non-fooled wrong answers: 45\%), and we saw no token-level hacks. But the judge also accepts unfinished answers: in steps 101--150 it accepted 97\% of the policy's lenient-wrong answers that were cut off at 256 tokens (P1.5$\times$Q1.5: 88\%), about as often as complete wrong answers ($99.97\%$ and $99.9\%$).

\begin{table}[ht]
\caption{\textbf{Error types of the 220 pre-registered fooled answers} (20 per cell; first error in reading order; coded by an LLM agent, one reader). Arith.: a stated computation is wrong. Misread: a number or relation from the problem is copied wrong or attached to the wrong object. Setup: the numbers are right but the plan is wrong (wrong operation, missing or extra step, or a different quantity than asked). Correct: the answer equals the reference, but the lenient parser misreads it. Ref.\ disp.: the answer follows a plain reading of the problem, but the GSM8K reference differs; the rest of the paper counts these as wrong. Other: one answer cut off at 256 tokens before its final answer (P0.5$\times$Q7; the generous label counts it as correct because its last line contains the reference number) and one copy slip (computes 1350, writes ``\texttt{\#\#\#\# 135}''). No \#\#\#\#: answers without ``\texttt{\#\#\#\# n}'' (not part of the row sum). Per-answer labels will be released with the code.}
\label{tab:errtypes}
\centering
\small
\setlength{\tabcolsep}{4pt}
\begin{tabular*}{\linewidth}{@{\extracolsep{\fill}}llrrrrrrr}
\toprule
Judge & Policy & Arith. & Misread & Setup & Correct & Ref.\ disp. & Other & No \#\#\#\# \\
\midrule
Q1.5 & 0.5B & 4 & 3 & 13 & 0 & 0 & 0 & 3 \\
Q1.5 & 1.5B & 6 & 2 & 12 & 0 & 0 & 0 & 3 \\
Q3 & 0.5B & 9 & 0 & 11 & 0 & 0 & 0 & 1 \\
Q3 & 1.5B & 3 & 1 & 15 & 0 & 1 & 0 & 0 \\
Q7 & 0.5B & 3 & 5 & 10 & 1 & 0 & 1 & 8 \\
Q7 & 1.5B & 6 & 1 & 5 & 4 & 3 & 1 & 7 \\
\midrule
L3 & 0.5B & 5 & 5 & 10 & 0 & 0 & 0 & 0 \\
L3 & 1.5B & 3 & 3 & 14 & 0 & 0 & 0 & 1 \\
\midrule
L8 & 0.5B & 8 & 2 & 10 & 0 & 0 & 0 & 1 \\
O1 & 0.5B & 4 & 3 & 13 & 0 & 0 & 0 & 1 \\
O7 & 0.5B & 3 & 3 & 14 & 0 & 0 & 0 & 4 \\
\midrule
All & & 54 & 28 & 127 & 5 & 4 & 2 & 29 \\
\bottomrule
\end{tabular*}
\end{table}

\begin{figure}[ht]
\centering
\includegraphics[width=0.87\linewidth]{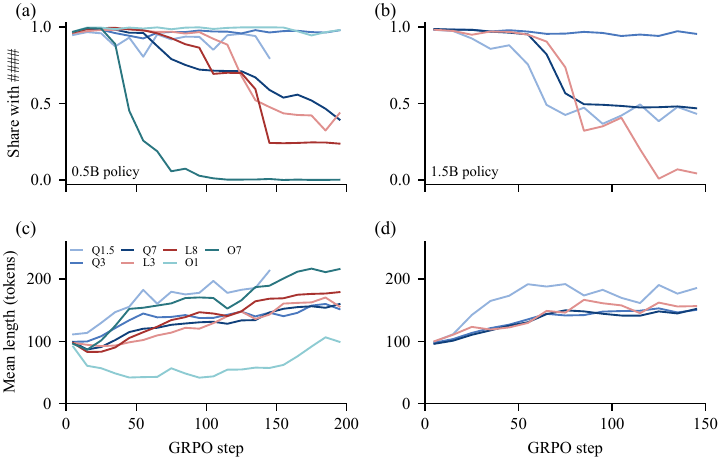}
\caption{\textbf{Answer format and length of the training rollouts} (the 11 cells of stages 1--5; the figure predates stage 6). Seed mean of 10-step window means. (a, b) Share of rollouts that contain ``\texttt{\#\#\#\# n}'', for the 0.5B and 1.5B policy. (c, d) Mean rollout length in tokens. Under Q7 with the 0.5B policy, only 2 of 4 seeds run past step 150; Q1.5 runs end at step 150.}
\label{fig:format}
\end{figure}

\textbf{Format drift.} Under L3, P1.5 rollouts with ``\texttt{\#\#\#\#}'' fall to 0.2\% and 29\% in steps 101--150, and two of four P0.5 seeds fall to 7.7\% and 0\% in steps 151--200; all eight Q3 runs keep the format (0.93--0.98). Under O7 all four seeds drop ``\texttt{\#\#\#\#}'' by steps 51--100 (toward \texttt{\textbackslash boxed\{n\}} or long prose), and under L8 three of four seeds drop it in steps 101--200. Under O1 the format stays, but accuracy collapses. In one L3 run, wrong answers in the drifted format are fooled more often than in the original format (0.95 vs.\ 0.81, post hoc). Figure~\ref{fig:format} shows the format share and the length of the training rollouts for the 11 cells of stages 1--5 (the stage-6 format share is in Section~\ref{sec:gap}). P1.5 under Q1.5 also drifts (share $0.44$ in steps 101--150, seed mean); this cell was not in the parser audit. Mean rollout length rises from 96--111 tokens in steps 1--10 to 135--191 tokens in steps 101--150 in every stage 1--5 cell except O1, where it falls from 91 to 54 tokens.

\textbf{Other onset definitions} (stage-2 fit, 95\% intervals for $\Delta_P$). Strict label: Q3 $+0.36$, Q7 $-0.46$ $[-0.92, -0.01]$. One common $\tau$ per judge: Q3 $+0.34$ $[-0.03, 0.70]$, Q7 $-3.0$ $[-\infty, -0.91]$. Threshold $0.7$: Q3 $+0.37$ $[0.10, 0.65]$, Q7 not identifiable (all censored).

\section{Example answers}
\label{app:examples}

We show four answers from the samples that we read (Appendix~\ref{app:extra}): (a), (c) and (d) here, and (b) in Figure~\ref{fig:fooled} (Section~\ref{sec:gold}). We chose them to show the typical cases; they are not a random sample. A random sample follows in (e). ``Accepted'' means $G_{\text{cal}}\geq\tau$ for that cell. Answers are verbatim, with line breaks kept.

\textbf{(a) A weak judge accepts a fluent answer with one arithmetic slip.}\par\nopagebreak[4]
\noindent\fbox{\parbox{0.97\linewidth}{\small\textbf{P0.5$\times$Q3, training step 22. Reference answer 12000. Model answer 1550. $G_{\text{cal}}=0.68$, accepted.}\\[2pt]
\textit{Question.} Sun City has 1000 more than twice as many people as Roseville City. Roseville city has 500 less than thrice as many people as Willowdale city. If Willowdale city has 2000 people, how many people live in Sun City?\\[2pt]
\textit{Answer.}\\
{\ttfamily\footnotesize If Willowdale city has 2000 people, then Riceville city has 3 * 2000 - 500 = 550 people.\\
Since Sun City has 1000 more people than Riceville city, it has 550 + 1000 = 1550 people.\\
\#\#\#\# 1550}}}
\vspace{6pt}

The answer makes two errors ($3\times2000-500=550$, and ``1000 more'' instead of ``1000 more than twice''), and it misnames the city. It is short and keeps the expected format.

\emph{The same policy at step 0.} The calibration set (Appendix~\ref{app:details}) holds two step-0 samples of P0.5 for this problem. Both are wrong. Q3 rejects the first and accepts the second ($\tau=0.24$ in this cell):

\noindent\fbox{\parbox{0.97\linewidth}{\small\textbf{P0.5, step 0, sample 1. Model answer 7000. $G_{\text{cal}}=0.15$, rejected.}\\[2pt]
{\ttfamily\footnotesize Roses city has 6000 people because 3 * 2000 + 500 = 6000\\
Sun city has 7000 people because 1000 + 6000 = 7000\\
\#\#\#\# 7000}\\[4pt]
\textbf{P0.5, step 0, sample 2. Model answer 14500. $G_{\text{cal}}=0.32$, accepted.}\\[2pt]
{\ttfamily\footnotesize Roseville City has 2*2000 + 1000 = 4000+1000 = 5000 people.\\
Sun City has 3*5000 - 500 = 15000 - 500 = 14500 people.\\
\#\#\#\# 14500}}}
\vspace{6pt}

The step-0 policy already writes short, fluent, wrong solutions to this problem, and the judge already accepts one of them. The step-22 answer in (a) gets a higher score ($0.68$).

\begin{table}[t]
\caption{\textbf{One random fooled answer per cell.} Step: training step. Model answer: the lenient parse. Error type: the first error, coded as in Table~\ref{tab:errtypes}, with a short note.}
\label{tab:randsample}
\centering
\small
\setlength{\tabcolsep}{4pt}
\begin{tabular}{llrrrrl}
\toprule
Judge & Policy & Step & Reference & Model answer & $G_{\text{cal}}$ & Error type \\
\midrule
Q1.5 & 0.5B & 30 & 39 & 30 & 0.70 & Setup (wrong pizza count, tip per pizza) \\
Q1.5 & 1.5B & 30 & 20 & 27 & 0.67 & Setup (berries grouped wrongly) \\
Q3 & 0.5B & 25 & 45 & 30 & 0.55 & Arith.\ (30\% of 150 $=$ 30) \\
Q3 & 1.5B & 61 & 164 & 54 & 0.45 & Setup ($\tfrac34$ of the total set equal to 41) \\
Q7 & 0.5B & 131 & 8 & 8.1 & 0.73 & Misread (4 ft 6 in read as 4.6 ft) \\
Q7 & 1.5B & 108 & 140 & 120 & 0.88 & Ref.\ disp.\ ($100+20=120$) \\
L3 & 0.5B & 36 & 8 & 5.8 & 0.52 & Setup (subtracts the 12 left from the 70 given) \\
L3 & 1.5B & 61 & 378 & 756 & 0.46 & Setup (doubles the total at the end) \\
L8 & 0.5B & 54 & 67 & 10 & 0.50 & Setup (episodes $\times$ 7 days per week) \\
O1 & 0.5B & 37 & 45 & 30 & 0.56 & Setup (600/5 left after eating 2/5) \\
O7 & 0.5B & 24 & 100 & 18 & 0.32 & Setup (3 worms per crow per hour) \\
Qw3-0.6 & 1.5B & 39 & 15 & 30 & 0.29 & Other (cut at 256 tokens; setup error) \\
\bottomrule
\end{tabular}
\end{table}

\textbf{(c) Format drift under the Llama judge.}\par\nopagebreak[4]
\noindent\fbox{\parbox{0.97\linewidth}{\small\textbf{P1.5$\times$L3, training step 64. Reference answer 6. Model answer 0. $G_{\text{cal}}=0.54$, accepted.}\\[2pt]
\textit{Question.} Bernie loves eating chocolate. He buys two chocolates every week at the local store. One chocolate costs him \$3. In a different store, there is a long-term promotion, during which each chocolate costs only \$2. How much would Bernie save in three weeks, if he would buy his chocolates in this store instead of his local one?\\[2pt]
\textit{Answer.}\\
{\ttfamily\footnotesize In a local store Bernie buys two chocolates every week and each costs \$3, meaning the total cost to him is 2*3=\$6\\
In the different store, the same number of chocolates costs 2*3=\$6\\
The savings Bernie gets by buying his chocolate in the long-term promotion store is 6-6=\$0\\
Bernie would be able to save 0 dollars per week if he buys his chocolates at the long-term promotion store\\
The answer is 0.}}}
\vspace{6pt}

The policy has dropped ``\texttt{\#\#\#\# n}'' for ``\texttt{The answer is n}'' (Appendix~\ref{app:extra}, format drift). The answer uses the wrong price in the second store.

\textbf{(d) A parser error: a correct answer read as wrong.}\par\nopagebreak[4]
\noindent\fbox{\parbox{0.97\linewidth}{\small\textbf{P1.5$\times$Q7, training step 111 (parser audit). Reference answer 240. Lenient parser reads 2. $G_{\text{cal}}=0.90$, accepted. Complete answer (106 tokens; blank lines removed).}\\[2pt]
\textit{Question.} Bob can shuck 10 oysters in 5 minutes. How many oysters can he shuck in 2 hours?\\[2pt]
\textit{Answer.}\\
{\ttfamily\footnotesize If Bob can shuck 10 oysters in 5 minutes, then he can shuck 10/5 = 2 oysters per minute.\\
There are 60 minutes in an hour, so in 2 hours, there are 2 * 60 = 120 minutes.\\
If Bob can shuck 2 oysters per minute for 120 minutes, then he can shuck 2 * 120 = 240 oysters in 2 hours.}}}
\vspace{6pt}

The answer is correct, but the last number in it is the ``2'' of ``2 hours''. All 17 parser errors in the audit have this form.

\textbf{(e) A random sample.} Examples (a)--(d) were chosen by hand. Table~\ref{tab:randsample} instead draws one answer per cell at random from the 20 pre-registered fooled answers (Python \texttt{random.Random(261011)}, one index from 0--19 per cell, cells in the order of the table; the stage-6 row continues the same sequence).

\section{Toward a predictive account}
\label{app:next}

The pre-registered test leaves step-0 AUC as the best simple predictor within one policy and shows that the logit scale does not transfer across families. A stronger version needs (i) both policies on the held-out judges, so that the pooled-row advantage seen in screening can be tested; (ii) at least three sizes in two or more new families; (iii) predictors that are invariant to the judge's logit scale; (iv) a judge smaller than the policy from the policy's own family, to separate size from family and template; and (v) a second task (a MATH subset). With the 0.5B policy, one task, 4 seeds and 200 steps, a new cell cost 2.5--4.4 A100-hours in stage 5.

\end{document}